\documentclass[sigconf, nonacm]{acmart}
\acmSubmissionID{2724} 

\usepackage{multirow}  
\usepackage{subcaption}
\usepackage[table,xcdraw]{xcolor}
\usepackage{graphicx}
\newcommand{\cmark}{\checkmark}

\renewcommand\footnotetextcopyrightpermission[1]{} 
\setcopyright{none}
\AtBeginDocument{%
  }

\begin{document}

\title{Mining Forgery Traces from Reconstruction Error: A Weakly Supervised Framework for Multimodal Deepfake Temporal Localization}

\author{Midou Guo}
\email{guomd5@mail2.sysu.edu.cn}
\orcid{0009-0002-0821-986X}
\affiliation{%
  \institution{School of Computer Science and Engineering \\Sun Yat-sen University}
  \city{Guangzhou}
  \country{China}
}

\author{Qilin Yin}
\email{yinqlin@mail2.sysu.edu.cn}
\orcid{0000-0001-7571-046X}
\affiliation{%
  \institution{School of Computer Science and Engineering \\Sun Yat-sen University}
  \city{Guangzhou}
  \country{China}
}

\author{Wei Lu}
\orcid{0000-0002-4068-1766}
\authornote{Corresponding author.}
\email{luwei3@mail.sysu.edu.cn}
\affiliation{
  \institution{School of Computer Science and Engineering \\Sun Yat-sen University}
  \city{Guangzhou}
  \country{China}
  }

\author{Rui Yang}
\email{duming.yr@alibaba-inc.com}
\affiliation{
  \institution{Alibaba Group}
  \city{Hangzhou}
  \country{China}
} 

\renewcommand{\shortauthors}{Midou Guo et al.}


\begin{abstract}
  Modern deepfakes have evolved into localized and intermittent manipulations that require fine-grained temporal localization to mitigate severe digital security risks. The prohibitive cost of frame-level annotation makes weakly supervised methods a practical necessity, which rely only on video-level labels. To this end, we propose Reconstruction-based Temporal Deepfake Localization (RT-DeepLoc), a weakly supervised temporal forgery localization framework that identifies forgeries via reconstruction errors. Our framework uses a Masked Autoencoder (MAE) trained exclusively on authentic data to learn its intrinsic spatiotemporal patterns; this allows the model to produce significant reconstruction discrepancies for forged segments, effectively providing the missing fine-grained cues for accurate localization without demanding dense human annotations. To robustly leverage these indicators, we introduce a novel Asymmetric Intra-video Contrastive Loss (AICL). By focusing on the compactness of authentic features guided by these reconstruction cues, AICL establishes a stable decision boundary that enhances local discrimination while preserving generalization to unseen forgeries by advanced generative models. Extensive experiments on large-scale datasets, including LAV-DF, demonstrate that RT-DeepLoc achieves state-of-the-art performance in weakly-supervised temporal forgery localization.
\end{abstract}




\maketitle

\section{Introduction}
With the rapid advancement of generative artificial intelligence technologies, highly realistic Deepfakes have been widely adopted across various digital platforms. However, their misuse poses serious risks to social stability and personal privacy. Therefore, developing effective Deepfake detection methods \cite{8630787, 8630761, li2021frequency, sun2021improving, he2021forgerynet, wang2023altfreezing, xu2023tall, sheng2025sumi, sheng2024dirloc, xia2024inspector, xia2024advancing, luo2023beyond, luo2024forgery, yin2023dynamic, yin2024fine}has become urgently needed for modern multimedia forensics.
\begin{figure}[t]
    \centering
    \includegraphics[width=\columnwidth]{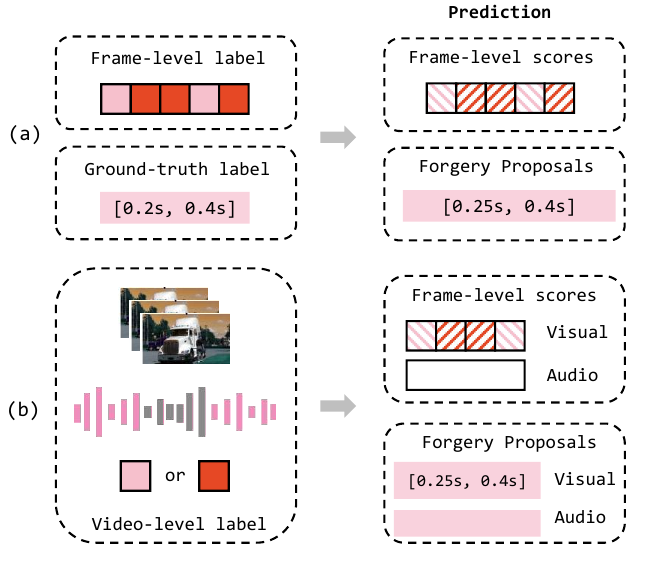}
    \caption{Comparison of different temporal forgery localization tasks: (a) Fully supervised temporal forgery localization; (b) Mutimodal weakly supervised temporal forgery localization.}
    \label{fig:Intro}
\end{figure}

Early deepfake detection research primarily focused on global detection\cite{guo2025towards, lv2024domainforensics, zhou2024fine}, treating the problem as a binary classification task to determine whether a multimedia sample is real or fake. However, modern deepfake techniques increasingly adopt localized and temporally intermittent manipulations. Forgers often tamper with only specific, critical segments of a video or audio stream (e.g., maliciously editing a few seconds of speech or facial expressions) while leaving the majority of the content distinctively authentic. In such partially manipulated scenarios, traditional global detectors are insufficient, as they fail to pinpoint the specific location of the forgery. Consequently, the research focus has necessarily shifted from coarse-grained classification to fine-grained temporal forgery localization, aiming to identify the precise timestamps of the manipulated fragments for more reliable multimedia forensics.

While temporal forgery localization is critical, the prohibitive cost of data annotation poses a core bottleneck in its practical application. Existing localization methods \cite{cai2022you, liu2023audio, zhang2023ummaformer} largely rely on fully supervised learning, as shown in Figure \ref{fig:Intro}(a). This paradigm requires precise, frame-level temporal annotations, entailing a labor-intensive process of meticulous human inspection that is impractical and costly to collect in real-world scenarios. To address this challenge, the researches have begun to explore weakly supervised learning (WSL) \cite{zhang2025cpl, wu2025weakly}. As illustrated in Figure \ref{fig:Intro}(b), WSL assumes only coarse-grained video-level labels (e.g., ``real'' or ``fake'') are available during training. Most existing WSL methods primarily construct a universal forgery feature boundary. These methods implicitly assume feature consistency among forged segments, attempting to distinguish them from authentic content via clustering. However, given the high heterogeneity of forgeries, they lack a unified distribution. Forcing these diverse patterns into a compact cluster distorts the feature space, impairing generalization against unseen attacks.

To address this weakness, we propose RT-DeepLoc, a novel weakly supervised temporal forgery localization framework that shifts the forgery forensics paradigm from modeling unbounded forgery patterns to learning the intrinsic spatiotemporal consistency inherent in authentic videos. The core of our framework is a forgery discovery network based on a Masked Autoencoder (MAE). By training this network exclusively on authentic videos to learn their intrinsic spatiotemporal regularities, it can effectively generate significant reconstruction errors when encountering forged segments, thereby providing strong forgery indicators without requiring frame-level labels. 
To effectively exploit these reconstruction-based cues, we further introduce a novel Asymmetric Intra-video Contrastive Loss (AICL). Unlike conventional weakly supervised methods that tend to force diverse forgery patterns into a single compact cluster, AICL leverages reconstruction errors to adaptively pinpoint ``forgery hotspots" through an asymmetric triplet mining strategy. Specifically, it encourages compact authentic feature representations while only separating them from the features of forgery hotspots. This targeted separation prevents the model from overfitting to specific forgery types and thus preserves the generalization of the feature space for unseen manipulations.
Finally, a Multi-task Learning Reinforcement strategy is explicitly introduced to bridge the semantic gap between discriminative original features and generative reconstructed cues. Built upon a synergistic dual-stream architecture with predictive consistency constraints, this strategy aligns the two parallel streams to fully exploit their complementary strengths. This explicit cross-stream supervision serves to regularize the complex learning process, mitigating predictive uncertainty and ultimately ensuring robust, stable, and more reliable localization predictions.

The main contributions of this work are summarized as follows:

\begin{itemize}
  \item We propose RT-DeepLoc, a novel weakly supervised framework that utilizes reconstruction errors as strong forgery indicators to achieve precise multimodal temporal localization with only video-level labels. 
  \item We propose a novel Forgery Discovery Network based on MAE, which captures intrinsic spatiotemporal consistency to amplify forgery traces into significant reconstruction errors, serving as robust unsupervised indicators.
  \item We introduce a novel Asymmetric Intra-video Contrastive Loss module that improves the compactness of authentic features and their separation from forgeries, enhancing local localization performance.
\end{itemize}

\section{Related Works}

\subsection{Weakly supervised Learning}
Weakly Supervised Learning (WSL) mitigates the bottleneck of prohibitive data annotation costs for tasks like temporal localization by relying exclusively on coarse, video-level labels (e.g., ``real'' or ``fake''). In traditional temporal action localization (TAL), the success of WSL methods is largely attributed to their ability to exploit the high intra-class consistency of semantic actions (e.g., ``running'' or ``jumping''), allowing them to identify common spatiotemporal patterns without precise frame-level guidance \cite{gao2022fine, wang2023weakly, wang2023temporal}. Inspired by these advancements, recent studies (e.g., CPL \cite{zhang2025cpl} and LOCO \cite{wu2025weakly}) have attempted to adapt WSL principles for temporal forgery localization (TFL). However, directly transplanting this paradigm into the deepfake domain encounters fundamental challenges. Unlike the high cohesiveness of semantic actions, forgery patterns exhibit extreme heterogeneity, which lacks a unified distribution, manifesting in diverse and unpredictable ways. Under standard weak supervision, driven purely by video-level binary classification objectives, models are forced to group these highly disparate and heterogeneous forged features into a single, compact cluster. This forced aggregation inevitably distorts the underlying feature space. Consequently, models become highly prone to overfitting to the specific manipulation types present in the training set, which severely impairs their generalization ability against unseen, novel forgery attacks. Our proposed RT-DeepLoc framework is specifically designed to overcome this bottleneck. Rather than attempting to model unbounded and infinitely diverse forgery patterns, we shift the paradigm toward capturing the intrinsic, stable spatiotemporal consistency inherent in authentic data. Through this fundamental paradigm shift, our model effectively isolates forgery traces as anomalies that disrupt these natural regularities, thereby achieving precise fine-grained temporal localization under weak supervision.


\begin{figure*}[ht]
    \centering
    \includegraphics[width=\textwidth]{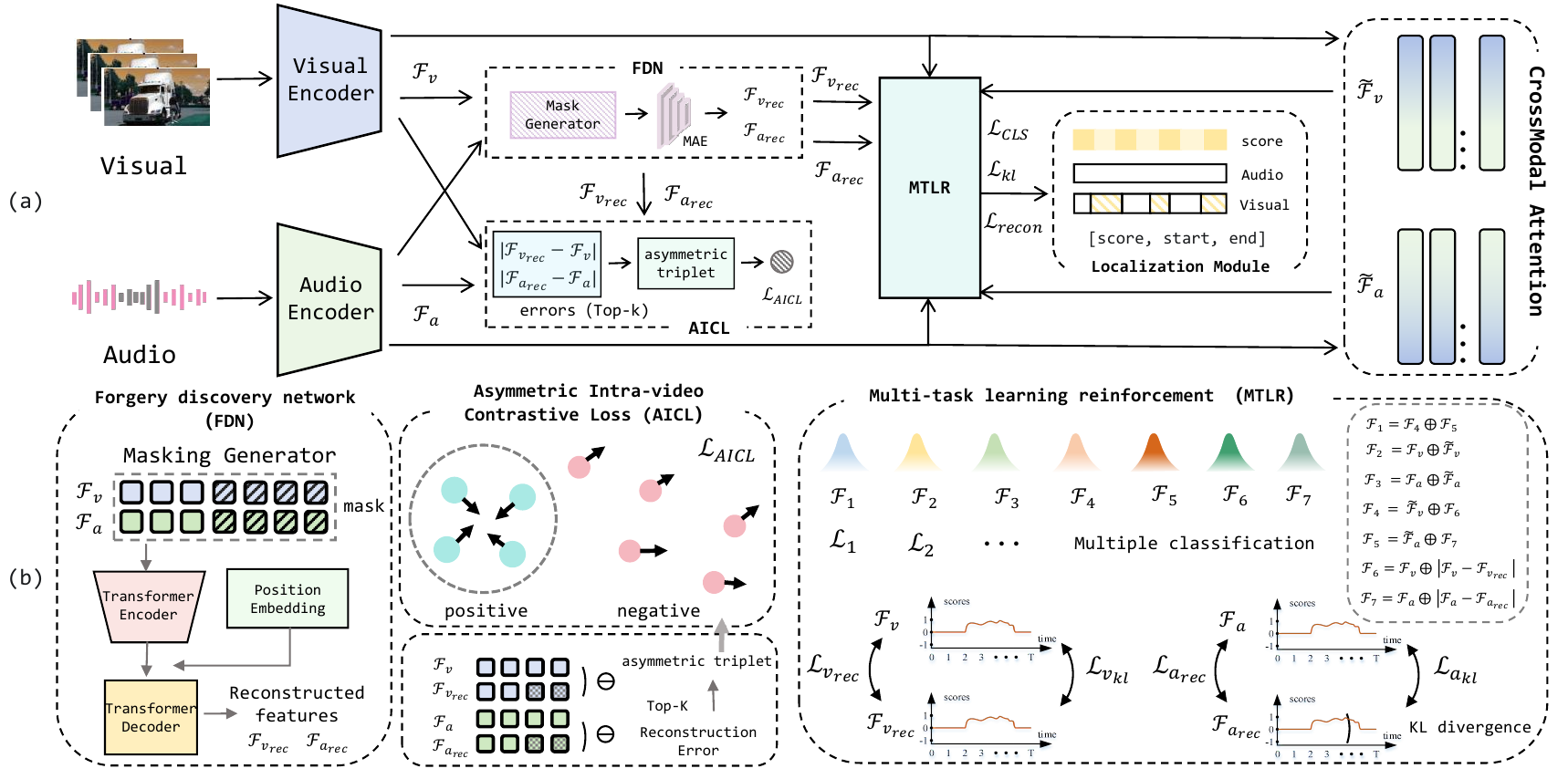}
    \caption{(a) The overall workflow and data flow of the proposed framework. (b) The internal architecture of the core components within RT-DeepLoc, which includes the Multimodal Feature Encoding and Fusion module, the Forgery Discovery Network based on MAE, the Asymmetric Intra-video Contrastive Loss module, and the Multi-task Learning Reinforcement strategy.}
    \label{fig:Main model}
\end{figure*}

\subsection{Mask Autoencoder} 
Originating from the image domain, MAE \cite{he2022masked} revolutionized self-supervised representation learning via a ``mask and reconstruct" paradigm. This success was subsequently extended to the audio/video domain with VideoMAE \cite{tong2022videomae, huang2022masked}. Distinct from traditional Autoencoders (AE) or Variational Autoencoders (VAE) that primarily focus on compressing and restoring global representations, VideoMAE can fully utilize the temporal correlation between video frames to reconstruct the masked areas from adjacent frames. This compels the model to learn robust temporal continuity. 
Given that modern deepfake manipulations (such as face-swapping or lip-syncing) inherently disrupt this delicate spatiotemporal consistency, they introduce imperceptible structural artifacts and temporal jitters. Consequently, a VideoMAE trained exclusively on authentic, unmanipulated data becomes an expert in the "intrinsic regularities" of real content. When encountering manipulated segments, the model struggles to accurately infer the forged patches from the surrounding authentic context, thereby yielding pronounced and localized reconstruction errors. These errors naturally serve as highly sensitive, unsupervised forgery indicators.

However, applying VideoMAE to weakly supervised multimodal deepfake temporal localization remains a challenging and unexplored direction. Direct application of standard VideoMAE is insufficient, as it lacks multimodal fusion and cannot effectively translate reconstruction errors into temporal localization. Our work fills this gap by redesigning VideoMAE into a dual-stream module that amplifies forgery traces to guide localization under weak supervision.

\section{Proposed Method}
\subsection{Overview}

This paper proposes RT-DeepLoc, a novel framework for weakly supervised multimodal deepfake temporal localization that mines forgery traces from reconstruction errors. We aim to precisely localize forged intervals in video and audio streams by relying exclusively on video-level 4-class labels as weak supervision. The overall framework is illustrated in Figure~\ref{fig:Main model}(a). The proposed framework consists of the Forgery Discovery Network (FDN), the Asymmetric Intra-video Contrastive Loss (AICL), and the Multi-task Learning Reinforcement strategy (MTLR). Among these components, the FDN is responsible for extracting robust reconstruction errors. Based on these errors, the AICL and MTLR reinforce the feature learning of the entire framework, enabling the transformation of coarse unsupervised cues into precise temporal boundaries using only video-level labels.

Specifically, visual and audio features are extracted via pre-trained backbones for their respective modalities and subsequently mapped by independent convolutional embedding layers into aligned temporal representations $\mathcal{F}_{v}$ and $\mathcal{F}_{a}$. These features are then fused using a bidirectional cross-modal attention mechanism, which facilitates mutual enhancement between modalities. By concatenating the original features with the enhanced outputs, the module generates a unified, contextually rich temporal representation $\mathcal{F}_{m}$ that serves as the foundation for subsequent processing.
Subsequently, the Forgery Discovery Network, which is built upon the Masked Autoencoder (MAE) framework, operates on the $\mathcal{F}_{v}$ and $\mathcal{F}_{a}$ to uncover anomalous cues, respectively. Specifically, a masking generator first obscures a high proportion (e.g., 75\%) of the input temporal sequences. Following the asymmetric design of MAE, a Transformer encoder processes only the sparse visible features, while a lightweight decoder attempts to reconstruct the full original sequence using the encoded representations and learnable mask tokens. To tailor this mechanism for forgery discovery, we implement a Genuine-Focused Reconstruction strategy, where the reconstruction loss $\mathcal{L}_{recon}$ is optimized exclusively on authentic samples. This constraint compels the network to internalize the intrinsic spatiotemporal regularities of real data, ensuring that any subsequent forgery manifests as a significant reconstruction discrepancy.

To leverage the reconstruction module's unsupervised indicators for fine-grained localization, we introduce the Asymmetric Intra-video Contrastive Loss $\mathcal{L}_{\text{AICL}}$. First, the discrepancy between original ($\mathcal{F}_{v}$, $\mathcal{F}_{a}$) and reconstructed features ($\mathcal{F}_{v_{rec}}$, $\mathcal{F}_{a_{rec}}$) is utilized to identify ``forgery hotspots" by selecting the top-K frame-level features exhibiting the largest reconstruction error. Based on these hotspots, we devise an asymmetric triplet mining strategy to construct (Anchor, Hardest Positive, Hardest Negative) triplets. A crucial aspect of this strategy is that anchors are exclusively sampled from authentic segments. This asymmetric approach is designed to prevent the model from incorrectly grouping features from diverse forgery methods into a single cluster, which would impair generalization. Ultimately, the $\mathcal{L}_{\text{AICL}}$ loss encourages authentic features to form a compact cluster while maximizing their distance from forged features, thereby enhancing frame-level forgery discrimination.

To synergize the discriminative original features ($\mathcal{F}_{v}$, $\mathcal{F}_{a}$) with their generative reconstructed counterparts ($\mathcal{F}_{v_{rec}}$, $\mathcal{F}_{a_{rec}}$), we employ a dual-branch architecture regularized by a cross-stream consistency constraint. By minimizing the Kullback-Leibler (KL) divergence between the predictive distributions of both branches, we compel the model to maintain stability across different feature views. 

During inference, the model dynamically routes the localization task to the most pertinent modality branch (visual, audio, or joint) based on the predicted video-level label. The final temporal boundaries are then determined by thresholding and merging the predicted frame-level scores.




\subsection{Forgery Discovery Network based on MAE}
\label{sec:MAE}
The core challenge in weakly supervised temporal forgery localization is detecting traces without fine-grained annotations. However, given the heterogeneity of forgery techniques, directly learning these unbounded patterns is challenging. Consequently, we shift the focus from modeling diverse forgeries to capturing the intrinsic consistency of authentic data. Accordingly, we design a Forgery Discovery Network (FDN) based on MAE  to reveal potential forged segments by reconstruction errors. The network is trained exclusively on authentic videos to learn stable spatiotemporal patterns. Consequently, forged segments disrupting these patterns yield significant reconstruction errors, serving as robust unsupervised localization indicators.

Specifically, as illustrated in Forgery Discovery Network (Figure \ref{fig:Main model}(b)), we denote the input temporal feature sequence $\mathbf{F} \in \mathbb{R}^{T \times C}$ for both modalities. To capture the intrinsic spatiotemporal regularities, we employ a random masking mechanism.  A subset of indices $\mathcal{M}$ is sampled to be masked, where the number of the masked set is governed by the masking ratio $\rho$ (i.e., $|\mathcal{M}| = \lfloor \rho T \rfloor$). The remaining unmasked indices constitute the visible set $\mathcal{V}$. A Transformer encoder $\mathcal{E}_{\text{enc}}$ then processes only the visible feature subset $\mathbf{F}_{\mathcal{V}}$ to generate its high-level semantic representation $\mathbf{Z}_{\mathcal{V}} = \mathcal{E}_{\text{enc}}(\mathbf{F}_{\mathcal{V}})$. For reconstruction phase, we initialize a full-length sequence $\mathbf{Z}_{\text{full}} \in \mathbb{R}^{T \times D_{\text{dec}}}$ for the Transformer decoder $\mathcal{E}_{\text{dec}}$. Each query embedding $\mathbf{z}_i$ within this sequence is conditioned on its visibility: for a visible frame ($i \in \mathcal{V}$), the embedding is formed by fusing the corresponding encoder output $\mathbf{Z}_{\mathcal{V}, \text{idx}(i)}$ with its sinusoidal position embedding $\mathbf{p}_i$; for a masked frame ($i \in \mathcal{M}$), we employ a shared, learnable mask token $\mathbf{e}_{\text{mask}}$ supplemented by $\mathbf{p}_i$. This embedding process can be formalized as:
\begin{equation}
    \mathbf{z}_i =
\begin{cases}
z_{\mathcal{V}, \text{idx}(i)} + \mathbf{p}_i & \text{if } i \in \mathcal{V} \\
e_{\text{mask}} + \mathbf{p}_i & \text{if } i \in \mathcal{M}
\end{cases}
\end{equation}
In this manner, the decoder $\mathcal{E}_{\text{dec}}$ operates on a complete sequence that encapsulates both the contextual cues from visible regions and the structural priors of the masked intervals. Consequently, it can predict the full reconstructed features $\hat{\mathbf{F}} = \mathcal{E}_{\text{dec}}(\mathbf{Z}_{\text{full}})$, which comprise $\mathcal{F}_{v_{rec}}$ and $\mathcal{F}_{a_{rec}}$ for the visual and audio modalities, respectively.

In this module, we employ the Genuine-Focused Reconstruction strategy, where the reconstruction loss $\mathcal{L}_{recon}$ is conditionally applied only to authentic samples. Under this constraint, the model effectively captures the intrinsic consistency of authentic data. Since forgeries inevitably violate these learned intrinsic patterns, the model struggles to reconstruct forged segments, resulting in significantly amplified reconstruction errors (calculated as $\left| \mathbf{F} - \hat{\mathbf{F}} \right|$), which serve as the key unsupervised indicators. Formally, the reconstruction loss $\mathcal{L}_{recon}$ is defined as:
\begin{equation}
    \mathcal{L}_{recon} = \frac{1}{|\mathcal{M}|} \sum_{j \in \mathcal{M}} \mathbb{I}(y = 0) \cdot \| \mathbf{F}_j - \hat{\mathbf{F}}_j \|_2^2
\end{equation}
where $\mathbf{F}_j$ and $\hat{\mathbf{F}}_j$ are the original and reconstructed features of the j-th masked frame, respectively, $\mathcal{M}$ represents the masked index set, and $\mathbb{I}(\cdot)$ is the indicator function that activates only for authentic videos ($y = 0$). This design trains the model to be an ``authentic content expert", making reconstruction errors a reliable indicator of forged regions.

\subsection{Asymmetric Intra-video Contrastive Loss}
While the Section \ref{sec:MAE} provides informative reconstruction cues, how to effectively utilize these local signals under weak supervision remains a hurdle. Naively clustering segments with high reconstruction errors overlooks the inherent heterogeneity of forgery patterns by implicitly assuming feature consistency. Such an assumption forces diverse forgery artifacts into compact clusters, inevitably distorting the feature space. To address this, we introduce a novel Asymmetric Intra-video Contrastive Loss (AICL). Rather than relying on global features, AICL utilizes reconstruction errors to adaptively identify ``forgery hotspots'' within each video, thereby concentrating the learning process on the most task-relevant local regions.


Specifically, as illustrated in Asymmetric Intra-video Contrastive Loss (Figure 2(b)), for each multimodal sample, we select the Top-K frames with the highest reconstruction difference score for each modal to form the set $\mathcal{K}$, respectively. We then extract the feature of these corresponding frames from the original feature sequence $\mathbf{F} \in \mathbb{R}^{T \times C}$ and generate a compact localized representative feature $\mathbf{f}_{\text{local}} = \frac{1}{K} \sum_{k \in \mathcal{K}} \mathbf{F}_k$. For the forged samples, these feature represents the most suspicious regions of the video. While for the authentic samples, these features correspond to the regions the model considers to be the most complex or distinctive. This local feature thus serves as the fundamental unit for our contrastive learning.

\begin{figure}[ht]
    \centering
    \includegraphics[width=0.8\columnwidth]{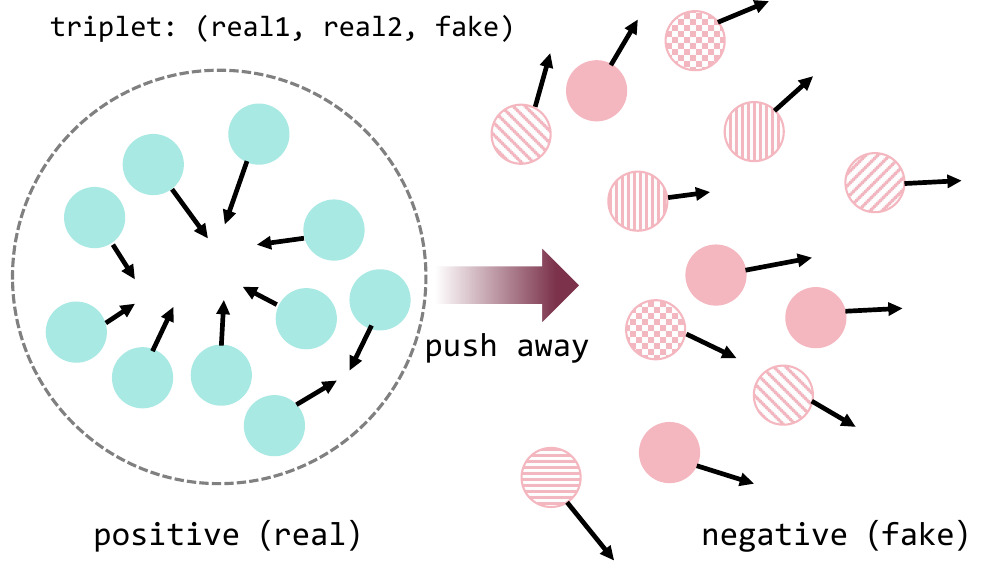}
    \caption{The AICL Mechanism. AICL enforces the compactness of authentic samples (blue circles) while pushing them away from diverse forgery features (varied pink circles).}
    \label{fig:AICL}
\end{figure}

Subsequently, we introduce an asymmetric triplet mining strategy to construct triplets, aiming to enhance the generalization capability of our framework. The core mechanism of this strategy is illustrated in Figure \ref{fig:AICL}. This approach is motivated by the diversity and rapid evolution of deepfake techniques, which result in heterogeneous forgery patterns, as depicted by the varied negative samples (pink circles) in Figure \ref{fig:AICL}. Forcing such heterogeneous forgeries into a single, compact cluster is not only impractical but also detrimental to generalization performance.
Therefore, our strategy focuses exclusively on structuring the authentic feature space.
Given a batch of samples, let $\mathcal{I}_{\text{real}} = \{ i \mid y_i = 0 \}$ denote the set of indices for authentic samples. We designate only the local features of these authentic samples ($\mathbf{f}_{\text{local}}^{(i)}$, $i \in \mathcal{I}_{\text{real}}$) as anchors. For each authentic feature $\mathbf{f}_{\text{local}}^{(i)}$, denoted as the anchor $\mathbf{f}_{\text{anc}}^{(i)}$, we perform hard sample mining to construct a triplet $\left( \mathbf{f}_{\text{anc}}^{(i)}, \mathbf{f}_{\text{pos}}^{(i)}, \mathbf{f}_{\text{neg}}^{(i)} \right)$ as follows:

\begin{itemize}
  \item \textbf{Hardest Positive} $\mathbf{f}_{\text{pos}}^{(i)}$:
  The local feature $\mathbf{f}_{\text{local}}^{(j)}$ of the authentic sample that is farthest from the anchor.
  \[
    \mathbf{f}_{\text{pos}}^{(i)} = \underset{\mathbf{f}_{\text{local}}^{(j)}: \, y_j=0, \, j \neq i}{\arg \max} \left\| \mathbf{f}_{\text{anc}}^{(i)} - \mathbf{f}_{\text{local}}^{(j)} \right\|_2
    \]

  \item \textbf{Hardest Negative} $\mathbf{f}_{\text{neg}}^{(i)}$: 
  The local feature $\mathbf{f}_{\text{local}}^{(k)}$ of the forged sample that is closest to the anchor.
    \[
    \mathbf{f}_{\text{neg}}^{(i)} = \underset{\mathbf{f}_{\text{local}}^{(k)}: \, y_k=1}{\arg \min} \left\| \mathbf{f}_{\text{anc}}^{(i)} - \mathbf{f}_{\text{local}}^{(k)} \right\|_2
    \]
\end{itemize}

This hard sample mining strategy provides the strongest learning signal by selecting the farthest positive sample to enforce intra-class compactness and the closest negative sample to improve inter-class separability. 

Based on the mined triplets, we compute $\mathcal{L}_{\text{AICL}}$ using the following triplet loss function:
\begin{align}
\mathcal{L}_{\text{AICL}}
&= \frac{1}{|\mathcal{I}_{\text{real}}|}
  \sum_{i \in \mathcal{I}_{\text{real}}}
  \max\Big(
      \lVert \mathbf{f}_{\text{anc}}^{(i)} - \mathbf{f}_{\text{pos}}^{(i)} \rVert_2^2
      \notag \\[-2pt]
&\qquad\qquad
      -
      \lVert \mathbf{f}_{\text{anc}}^{(i)} - \mathbf{f}_{\text{neg}}^{(i)} \rVert_2^2
      + m,\;
      0
  \Big)
\end{align}
where $\mathcal{I}_{\text{real}}$ denotes the index set of authentic samples within the batch, and $m$ is the margin hyperparameter. The terms $\mathbf{f}_{\text{anc}}^{(i)}$, $\mathbf{f}_{\text{pos}}^{(i)}$ and $\mathbf{f}_{\text{neg}}^{(i)}$ represent the anchor, hard positive, and hard negative features for the $i$-th authentic sample, respectively.

By minimizing this loss, the model learns a feature space where authentic features are compactly clustered and clearly separated from forged ones, greatly enhancing its discriminative ability for local forged regions.

\subsection{Multi-task Learning Reinforcement}
To fully exploit the synergy between the discriminative original features and the generative reconstructed features, we employ a Multi-task Learning Reinforcement (MTLR) strategy. The core of this strategy is to enforce predictive consistency across the dual streams, ensuring that the model's judgments remain stable regardless of the feature source.

Specifically, we introduce a probability distribution consistency loss, denoted as $\mathcal{L}_{\text{KL}}$. Since the original and reconstructed streams capture complementary views of the data, their predictive probability distributions should align on the final decision. We utilize the Kullback-Leibler (KL) divergence to penalize the discrepancy between the prediction probabilities of the original stream and the reconstructed stream. This alignment acts as a regularization mechanism, preventing the branches from diverging and ensuring that the reconstruction process retains sufficient discriminative semantics.

Complementing this consistency constraint, we impose comprehensive supervision through a multi-head classification objective. To bridge the gap between frame-level scores and video-level weak labels, we adopt a Top-k Multiple Instance Learning aggregation strategy. For a frame-level score sequence $\mathbf{S} = \{ \mathbf{s}_1, \mathbf{s}_2, \dots, \mathbf{s}_T \}$, we select the $K$ frames with the highest scores and aggregate them via average pooling to compute the video-level score vector $\mathbf{s}_{\text{video}} = \frac{1}{K} \sum_{i=1}^{K} \mathbf{s}_{\text{top-K}}^{(i)}$. This strategy is applied to seven classification heads ($\mathcal{L}_{\text{CLS}}$), supervising the fused main branch, the modality-separated branches, and the individual original/reconstructed streams. The specific feature combinations for each head are illustrated in Multi-task learning reinforcement (Figure \ref{fig:Main model}(b)), where $\oplus$ denotes the concatenation operation. This multi-dimensional supervision ensures that every component of the network contributes effectively to the forgery detection task. 

Finally, the overall framework is jointly optimized by minimizing a weighted sum of all loss terms:
\begin{equation}
    \mathcal{L}_{\text{total}}
    = \lambda_1 \mathcal{L}_{\text{CLS}}
    + \lambda_2 \mathcal{L}_{\text{recon}}
    + \lambda_3 \mathcal{L}_{\text{KL}}
    + \lambda_4 \mathcal{L}_{\text{AICL}}
\end{equation}
where $\mathcal{L}_{\text{recon}}$ represents the MAE reconstruction loss, $\mathcal{L}_{\text{AICL}}$ is the asymmetric contrastive loss, and $\lambda_i$ are hyperparameters balancing the contribution of each task.

\section{Experiments}
\subsection{Experimental Settings}
\textbf{Datasets.} To evaluate the effectiveness of the method, we conducted experiments on two large-scale multimodal Deepfake datasets: LAV-DF \cite{cai2023glitch} and AV-Deepfake1M \cite{cai2024av}. LAV-DF is a comprehensive multimodal dataset designed to evaluate audio-visual consistency, consisting of 36,431 authentic videos and 99,873 forged videos. It contains a diverse range of manipulation types, including face swapping and lip-syncing, providing a challenging benchmark for detecting cross-modal inconsistencies. AV-Deepfake1M is a large-scale dataset designed to test detection performance in open-world scenarios. It comprises over 1.14 million video clips, with approximately 420,000 real videos and 720,000 fake videos.

\textbf{Evaluation Metrics.} The mean Average Precision (mAP) and Average Recall (AR) are utilized as the evaluation metrics following standard protocols. The IoU thresholds of mAP are set as $[0.1 : 0.1 : 0.7]$, and the number of proposals is set as $20, 10, 5$ and $2$, respectively.

\subsection{Implementation Details} 
\textbf{Data Preprocessing.} We utilize pre-trained TSN \cite{wang2018temporal} and Wav2Vec \cite{baevski2020wav2vec} to extract visual and audio features, respectively. To address temporal resolution discrepancies, we apply temporal pooling to align audio features with visual frames, facilitating effective cross-modal interaction. 

\textbf{Training Settings.} Our framework is implemented using PyTorch and trained on a single NVIDIA GeForce RTX 4090 GPU. The model is optimized with a batch size of 32 and an initial learning rate of $1 \times 10^{-5}$. We set the mask ratio of the FDN to 75\% and the Top-k parameter for the AICL is set to 10. The loss term weights are set as $\lambda_2 = \lambda_3 = \lambda_4 = 0.1$, and for the seven-head $\mathcal{L}_{\text{CLS}}$, we assign a weight of 0.8 to the main fused branch and 0.1 to the six auxiliary branches.

\begin{table*}[h]
\centering
\caption[LAV-DF Results]{
    Performance comparison of temporal forgery localization methods on the LAV-DF dataset. For weakly supervised approaches, the best and second-best average AP and AR are highlighted in \textcolor{red}{\textbf{red}} and \textcolor{blue}{\textbf{blue}}, respectively.
}
\label{tab:results_lavdf}
\resizebox{\textwidth}{!}{%
    \begin{tabular}{c|c|cccccccc|ccccc}
    \hline
    \multirow{2}{*}{Supervision} & \multirow{2}{*}{Method} & \multicolumn{8}{c|}{mAP@IoU(\%)} & \multicolumn{5}{c}{AR@Proposals(\%)} \\ 
    & & 0.1 & 0.2 & 0.3 & 0.4 & 0.5 & 0.6 & 0.7 & Avg. & 20 & 10 & 5 & 2 & Avg. \\ \hline
    
    \multirow{4}{*}{fully} 
     & ActionFormer \cite{zhang2022actionformer} & 97.97 & 97.69 & 97.27 & 96.78 & 96.28 & 95.51 & 94.61 & 96.59 & 99.17 & 99.02 & 98.41 & 95.95 & 98.14 \\
     & TriDet \cite{shi2023tridet} & 94.99 & 94.74 & 94.35 & 93.83 & 93.19 & 92.20 & 90.67 & 93.42 & 97.12 & 96.93 & 96.31 & 93.68 & 96.01 \\
     & UMMAFormer \cite{zhang2023ummaformer} & 97.69 & 97.57 & 97.37 & 97.11 & 96.70 & 95.96 & 94.90 & 96.76 & 98.63 & 98.53 & 98.22 & 95.14 & 97.63 \\
     & MFMS \cite{zhang2024mfms} & 98.00 & 97.91 & 97.78 & 97.63 & 97.31 & 96.69 & 95.79 & 97.30 & 98.94 & 98.86 & 98.62 & 95.61 & 98.01 \\ \cline{1-15}
    
    \multirow{5}{*}{weakly} 
     & CoLA \cite{zhang2021cola} & 31.25 & 25.93 & 19.42 & 13.35 & 8.59 & 5.23 & 2.71 & 15.21 & 41.19 & 41.18 & 40.84 & 37.79 & 40.25 \\
     & FuSTAL \cite{feng2025full} & 31.55 & 25.40 & 19.16 & 13.48 & 8.91 & 5.58 & 2.95 & 15.29 & 39.07 & 39.05 & 38.75 & 36.09 & 38.24 \\
     & SAL \cite{li2025multilevel} & 12.72 & 3.86 & 1.78 & 1.04 & 0.58 & 0.26 & 0.09 & 2.90 & 15.56 & 15.53 & 15.47 & 14.55 & 15.28 \\
     & LOCO \cite{wu2025weakly} & 62.40 & 55.09 & 50.78 & 45.18 & 36.84 & 31.65 & 28.02 & \textcolor{blue}{\textbf{44.28}} & 52.31 & 52.31 & 52.31 & 52.19 & \textcolor{blue}{\textbf{52.28}} \\
     & \cellcolor{gray!15}\textbf{RT-DeepLoc} & \cellcolor{gray!15}90.14 & \cellcolor{gray!15}86.61 & \cellcolor{gray!15}81.06 & \cellcolor{gray!15}75.29 & \cellcolor{gray!15}67.43 & \cellcolor{gray!15}59.26 & \cellcolor{gray!15}50.31 & \cellcolor{gray!15}\textcolor{red}{\textbf{72.87}} & \cellcolor{gray!15}84.18 & \cellcolor{gray!15}84.18 & \cellcolor{gray!15}84.18 & \cellcolor{gray!15}83.61 & \cellcolor{gray!15}\textcolor{red}{\textbf{84.03}} \\ \hline
    \end{tabular}%
}
\end{table*}

\begin{table*}[h]
\centering
\caption[AV-Deepfake1M Results]{
    Temporal forgery localization results on the \textbf{AV-Deepfake1M} dataset. 
    Best results among weakly supervised methods are in \textcolor{red}{\textbf{red}}, second-best are in \textcolor{blue}{\textbf{blue}}.
}
\label{tab:results_avdeepfake1m}
\resizebox{\textwidth}{!}{%
    \begin{tabular}{c|c|cccccccc|ccccc}
    \hline
    \multirow{2}{*}{Supervision} & \multirow{2}{*}{Method} & \multicolumn{8}{c|}{mAP@IoU(\%)} & \multicolumn{5}{c}{AR@Proposals(\%)} \\ 
    & & 0.1 & 0.2 & 0.3 & 0.4 & 0.5 & 0.6 & 0.7 & Avg. & 20 & 10 & 5 & 2 & Avg. \\ \hline
    
    \multirow{4}{*}{fully} 
     & ActionFormer \cite{zhang2022actionformer} & 67.30 & 67.27 & 67.24 & 67.16 & 67.02 & 66.65 & 65.45 & 66.87 & 83.09 & 82.91 & 82.56 & 78.70 & 81.82 \\
     & TriDet \cite{shi2023tridet} & 55.68 & 55.60 & 55.43 & 55.14 & 54.69 & 53.82 & 51.73 & 54.58 & 74.89 & 74.17 & 72.85 & 67.95 & 72.47 \\
     & UMMAFormer \cite{zhang2023ummaformer} & 91.76 & 91.67 & 91.51 & 91.28 & 90.97 & 90.40 & 88.99 & 90.94 & 95.01 & 94.59 & 93.90 & 89.54 & 93.26 \\
     & MFMS \cite{zhang2024mfms} & 94.67 & 94.63 & 94.58 & 94.48 & 94.32 & 93.92 & 92.53 & 94.16 & 96.69 & 96.43 & 95.98 & 91.97 & 95.27 \\ \cline{1-15}
    
    \multirow{4}{*}{weakly} 
     & CoLA \cite{zhang2021cola} & 3.22 & 1.09 & 0.39 & 0.14 & 0.05 & 0.02 & 0.01 & \textcolor{blue}{\textbf{0.70}} & 20.71 & 19.95 & 16.13 & 8.17 & \textcolor{blue}{\textbf{16.24}} \\
     & FuSTAL \cite{feng2025full} & 3.03 & 1.02 & 0.40 & 0.15 & 0.05 & 0.02 & 0.01 & 0.67 & 19.67 & 18.62 & 14.55 & 7.08 & 14.98 \\
     & LOCO \cite{wu2025weakly} & 1.25 & 0.30 & 0.10 & 0.03 & 0.01 & 0.00 & 0.00 & 0.24 & 10.40 & 10.11 & 8.48 & 3.99 & 8.25 \\
     & \cellcolor{gray!15}\textbf{RT-DeepLoc} & \cellcolor{gray!15}48.94 & \cellcolor{gray!15}45.21 & \cellcolor{gray!15}41.94 & \cellcolor{gray!15}37.87 & \cellcolor{gray!15}29.41 & \cellcolor{gray!15}19.20 & \cellcolor{gray!15}7.68 & \cellcolor{gray!15}\textcolor{red}{\textbf{32.89}} & \cellcolor{gray!15}49.00 & \cellcolor{gray!15}48.95 & \cellcolor{gray!15}48.62 & \cellcolor{gray!15}47.03 & \cellcolor{gray!15}\textcolor{red}{\textbf{48.40}} \\ \hline
    \end{tabular}%
}
\end{table*}

\begin{table*}[h]
\centering
\caption{Generalization performance of cross-dataset evaluation.
}
\label{tab:cross-datasets}
\resizebox{\textwidth}{!}{%
    \begin{tabular}{c|c|cccccccc|ccccc}
    \hline
    \multirow{2}{*}{Supervision} & \multirow{2}{*}{Method} & \multicolumn{8}{c|}{mAP@IoU(\%)} & \multicolumn{5}{c}{AR@Proposals(\%)} \\ 
     & & 0.1 & 0.2 & 0.3 & 0.4 & 0.5 & 0.6 & 0.7 & Avg. & 20 & 10 & 5 & 2 & Avg. \\ \hline
    
    \multirow{2}{*}{fully} 
     & UMMAFormer \cite{zhang2023ummaformer} & 13.93 & 13.56 & 13.13 & 12.76 & 12.42 & 12.14 & 11.83 & 12.82 & 32.35 & 31.98 & 31.54 & 30.43 & 31.58 \\
     & MFMS \cite{zhang2024mfms} & 13.48 & 12.66 & 12.01 & 11.52 & 11.04 & 10.59 & 10.06 & 11.62 & 35.38 & 32.86 & 30.51 & 27.49 & 31.56 \\ \hline
    
    \multirow{3}{*}{weakly} 
     & CoLA \cite{zhang2021cola} & 1.35 & 0.13 & 0.03 & 0.01 & 0.01 & 0.01 & 0.01 & \textcolor{blue}{\textbf{0.22}} & 20.71 & 19.95 & 16.13 & 8.17 & \textcolor{blue}{\textbf{11.06}} \\
     & LOCO \cite{wu2025weakly} & 0.27 & 0.04 & 0.01 & 0.00 & 0.00 & 0.00 & 0.00 & 0.05 & 5.95 & 5.95 & 5.72 & 3.54 & 5.29 \\
     
    & \cellcolor{gray!15}\textbf{RT-DeepLoc} & \cellcolor{gray!15}21.81 & \cellcolor{gray!15}20.36 & \cellcolor{gray!15}18.39 & \cellcolor{gray!15}16.63 & \cellcolor{gray!15}14.88 & \cellcolor{gray!15}13.47 & \cellcolor{gray!15}11.06 & \cellcolor{gray!15}\textcolor{red}{\textbf{16.66}} & \cellcolor{gray!15}82.41 & \cellcolor{gray!15}82.41 & \cellcolor{gray!15}82.39 & \cellcolor{gray!15}77.55 & \cellcolor{gray!15}\textcolor{red}{\textbf{81.19}} \\ \hline
    \end{tabular}%
}
\end{table*}

\begin{table}[t]
\centering
\caption{Ablation study of different components in RT-DeepLoc on the LAV-DF dataset.}
\label{tab:ablation_wo}
\setlength{\tabcolsep}{12pt} 

\resizebox{1.0\columnwidth}{!}{%
    \begin{tabular}{ccc|cc}
    \toprule
    \textbf{FDN} & \textbf{MTLR} & \textbf{AICL} & \textbf{Avg. AP (\%)} & \textbf{Avg. AR (\%)} \\ 
    \midrule
    
     &  &  & 62.26 & 81.75 \\ 
    
    \cmark & & \cmark & 68.03 & 82.71 \\ 
    
    \cmark & \cmark & & 69.70 & 83.72 \\ 
    
    \midrule
    
    \rowcolor{gray!15} 
    \cmark & \cmark & \cmark & \textbf{72.87} & \textbf{84.03} \\ 
    \bottomrule
    \end{tabular}
}
\end{table}

\subsection{Intra-Dataset Evaluation}
In this section, we will compare the proposed RT-DeepLoc with the state-of-the-art methods previously used on LAV-DF and AV-Deepfake1M. 

\textbf{LAV-DF Datasets: } Table \ref{tab:results_lavdf} details the performance comparison on the LAV-DF dataset, where RT-DeepLoc establishes a commanding lead among weakly supervised approaches \cite{zhang2021cola, feng2025full, li2025multilevel, wu2025weakly}. Our method achieves a remarkable average AR of 84.03\%, surpassing the previous method (e.g., LOCO \cite{wu2025weakly}) by 31.75\%. This high recall serves as strong evidence that the reconstruction error indicators from FDN successfully act as sensitive ``forgery indicators,'' capturing subtle manipulation traces missed by other methods. Notably, the consistent performance from AR@2 to AR@20 demonstrates that RT-DeepLoc can predict forgery segments using a minimal number of proposals. Furthermore, the significant boost in mAP confirms that AICL effectively separates forged segments from authentic ones in the feature space. This clear separation enables the model to delineate precise temporal boundaries, overcoming the localization ambiguity typical of weak supervision. It is within expectation that fully-supervised methods \cite{zhang2022actionformer, shi2023tridet, zhang2023ummaformer, zhang2024mfms} maintain a performance lead, as they benefit from precise frame-level annotations that provide an ideal upper bound. However, compared to prior weakly-supervised works, RT-DeepLoc significantly narrows the performance gap between weak and full supervision. Overall, relying solely on coarse video-level labels, RT-DeepLoc effectively identifies temporal forgery traces within multi-modal features to achieve precise localization, offering a highly competitive and cost-effective alternative to fully-supervised paradigms.

\textbf{AV-Deepfake1M Datasets: }
The Table \ref{tab:results_avdeepfake1m} presents the results on the AV-Deepfake1M dataset. Compared to LAV-DF (7.6\% average forgery rate), AV-Deepfake1M exhibits a significantly lower forgery rate of only 3.6\%. This extremely low forgery rate, combined with the massive scale and high diversity poses a far more rigorous challenge. Experimental data reveals a catastrophic performance collapse in existing weakly-supervised methods. Both CoLA \cite{zhang2021cola} and LOCO \cite{wu2025weakly} fail to yield meaningful results, achieving negligible average APs of 0.70\% and 0.24\%, respectively. In sharp contrast, RT-DeepLoc demonstrates remarkable resilience, achieving an average AP of 32.89\% and an AR of 48.40\%. Although a gap with fully-supervised methods remains, our method outperforms existing weakly-supervised baselines by a significant margin. This result confirms that modeling the intrinsic consistency of authentic data is a strictly more robust paradigm. Even in such complex scenarios, reconstruction errors prove to be an effective and reliable localization signal. 

%

    
    
    
    

\begin{figure*}[t]
    \centering
    \includegraphics[width=\textwidth]{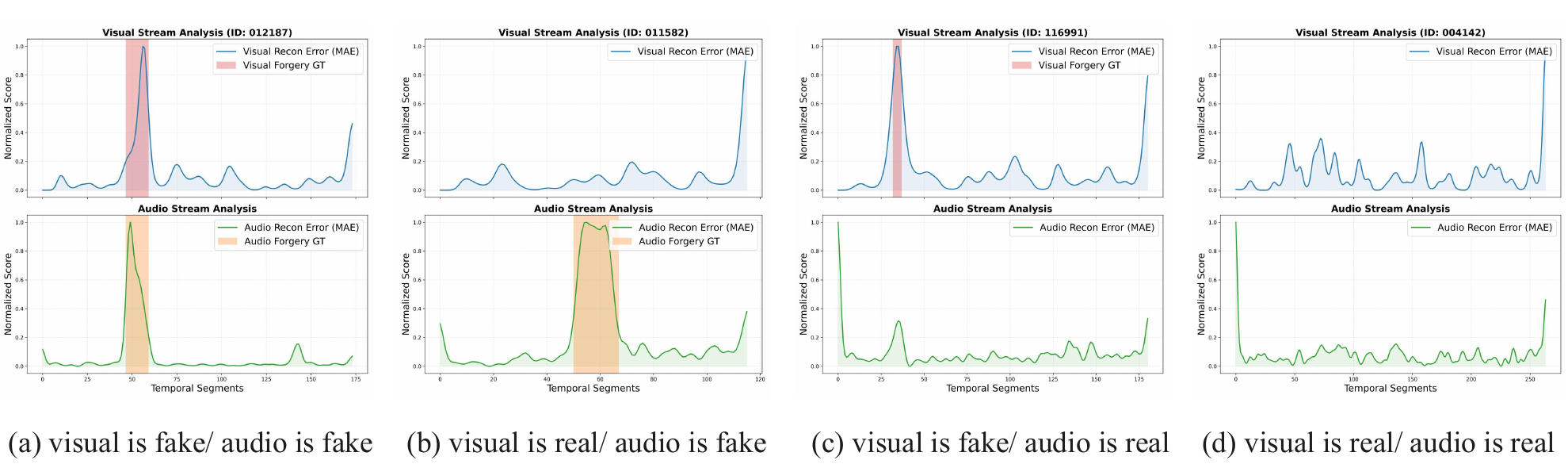}
    \caption{Qualitative visualization of modality-specific reconstruction discrepancies on LAV-DF. We present four scenarios: (a) audio-only, (b) multimodal, (c) visual-only forgeries, and (d) authentic video. Blue and green curves represent visual and audio reconstruction errors, respectively, while shaded areas indicate ground-truth intervals.}
    \label{fig:Qualitative visualization of MAE reconstruction discrepancy}
\end{figure*}

    
    
    
    

\subsection{Cross-Dataset Evaluation}
To evaluate the generalization ability of the proposed framework, we conducted cross-dataset experiments: training on AV-Deepfake1M and testing directly on LAV-DF without fine-tuning. Table \ref{tab:cross-datasets} details the assessment results. The results clearly demonstrate that the weakly-supervised method \cite{zhang2021cola, wu2025weakly} is completely ineffective, and even the fully supervised methods \cite{zhang2023ummaformer, zhang2024mfms} have experienced a sharp performance decline, achieving average APs of 12.82\% and 11.62\% respectively. This indicates that existing methods relying on learning to forge boundaries tend to overfit the forged patterns of the source domain and fail to generalize when the forgery distribution shifts. However, the RT-DeepLoc we proposed demonstrated extraordinary robustness and adaptability, achieving an average AP of 16.66\% and an average AR as high as 81.19\% respectively. The most significant finding from this experiment is that our weakly-supervised framework decisively surpasses the fully supervised methods in this unseen domain. This confirms that the intrinsic consistency of authentic data is a universal attribute shared across datasets. When traditional discriminative cues become unreliable due to domain offset, the reconstruction errors remain a robust indicator. This enables RT-DeepLoc to effectively detect inconsistencies in unseen data distributions, proving that modeling the bounded space of authentic data is a fundamentally more generalizable paradigm than learning the unbounded space of forgery.

\subsection{Ablation}
\subsubsection{Components Analysis}
To verify the contribution of each component in RT-DeepLoc, we conducted ablation studies on the LAV-DF dataset by removing the Forgery Discovery Network (FDN), the Multi-task Learning Reinforcement (MTLR), and the Asymmetric Intra-video Contrastive Loss (AICL). The results are summarized in Table \ref{tab:ablation_wo}. 

We first analyze the w/o FDN variant. Since both AICL and MTLR rely on reconstruction indicators, removing FDN inherently disables them. Consequently, the model reverts to a baseline that retains only multimodal encoders and crossmodal attention. Remarkably, this baseline achieves 62.26\% Avg. AP, significantly outperforming LOCO (44.28\%) by capturing semantic inconsistencies via attention. However, it still trails the full RT-DeepLoc by 10.61\%. This gap confirms that semantic cues alone are insufficient and FDN provides indispensable reconstruction indicators to uncover low-level artifacts and intrinsic anomalies that the baseline overlooks. The ablation of AICL results in a 3.17\% decrease in Avg. AP, while the Avg. AR remains relatively stable. This phenomenon perfectly aligns with our design motivation: while FDN is responsible for discovering potential anomalies, AICL focuses on refining the feature boundaries in the metric space. By enforcing asymmetric compactness, AICL effectively suppresses noise and sharpens the temporal localization, thereby significantly boosting mAP. Finally, removing the MTLR strategy leads to a degradation of 4.84\% in mAP. The MTLR module, by enforcing predictive consistency via KL divergence and multi-head supervision, effectively regularizes the dual-stream learning process, ensuring more robust and confident predictions. In summary, the full RT-DeepLoc framework achieves the best performance, demonstrating that these three components are complementary and collectively essential for precise weakly-supervised localization.

\subsubsection{Sensitivity Analysis of Hyperparameters}
\begin{figure}[t]
    \centering    
    \includegraphics[width=\columnwidth]{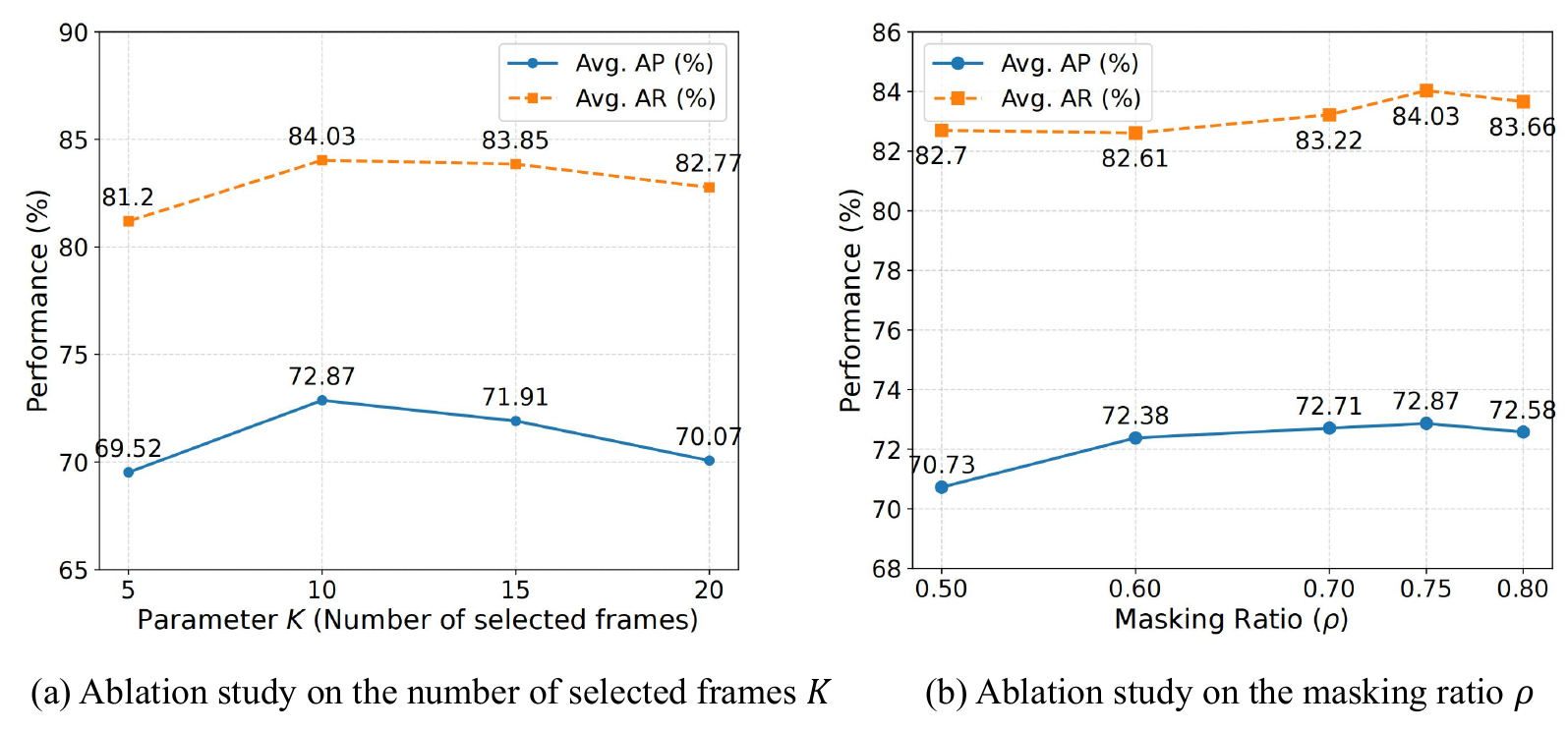}
    \caption{Sensitivity analysis of hyperparameters on the LAV-DF dataset. (a) The effect of the number of selected frames $K$ in the \textbf{AICL} module. (b) The effect of the masking ratio $\rho$ in the \textbf{FDN} module.}
    \label{fig:ablation_k_rho}
\end{figure}

\textbf{Parameter $K$ in the \textbf{AICL} module: } 
We further investigate the impact of the hyperparameter $K$, the number of ``forgery hotspots'' frames selected for AICL. As shown in Figure \ref{fig:ablation_k_rho}(a), the model achieves optimal performance at $K=10$ (72.87\% Avg. AP and 84.03\% Avg. AR). Reducing $K$ to 5 leads to a performance drop, because an overly sparse selection fails to cover forgery patterns to represent the diverse manipulation distribution. Conversely, increasing $K$ to 15 or 20 also degrades performance. A larger $K$ introduces noise by misidentifying authentic frames as negatives. This label noise confuses the asymmetric contrastive learning process, impairing the discriminative power of the learned features. Therefore, we adopt $K=10$ as the default setting to balance between signal sufficiency and purity.

\textbf{Parameter $\rho$ in the \textbf{FDN} module: } 
We investigate the sensitivity of FDN to the masking ratio $\rho$, varying it from 0.50 to 0.80. As shown in Figure \ref{fig:ablation_k_rho}(b), the performance initially improves with the increase of $\rho$, peaking at $\rho=0.75$ (72.87\% Avg. AP, 84.03\% Avg. AR). When $\rho$ is relatively low (e.g., 0.50), the reconstruction task becomes trivial; abundant visible patches allow reconstruction via simple interpolation without capturing intrinsic consistency. Consequently, the model can easily reconstruct even forged segments, resulting in indistinguishable error signals. Conversely, a high ratio (e.g., 0.80) removes too much contextual information, introducing noise into the reconstruction error maps, leading to a slight performance degradation. Therefore, we set $\rho=0.75$ to balance reconstruction difficulty for robust learning with sufficient context for inference.

\subsubsection{Qualitative Analysis}
To intuitively validate the effectiveness of the proposed FDN, we visualize the frame-level reconstruction errors across different scenarios in Figure \ref{fig:Qualitative visualization of MAE reconstruction discrepancy}. These four scenarios correspond to: (a) audio-only forgery; (b) multimodal forgery; (c) visual-only forgery; and (d) authentic video. The horizontal axis represents the temporal sequence of video segments, while the vertical axis denotes the normalized reconstruction error. The blue and green curves denote visual and audio errors, respectively, with shaded regions marking ground-truth forgeries.

As illustrated, the reconstruction error exhibits a strong correlation with the presence of forgeries. In the manipulated segments (shaded regions in a-c), we observe distinct spikes in the corresponding error curves. These surges align precisely with the ground-truth intervals, indicating that the FDN successfully amplifies the deviations caused by forgery traces. It is worth noting that a minor increase in reconstruction error is observed at the video boundaries (e.g., the end of the timeline, which is expected. Since the MAE relies on bidirectional context to infer masked features, the scarcity of contextual information at the sequence edges inevitably leads to slightly higher reconstruction uncertainty. 
However, compared to the sharp peaks from actual forgeries, these boundary effects are negligible and do not harm localization accuracy.

In summary, this visualization confirms our core premise: forgeries disrupt spatiotemporal consistency, making reconstruction discrepancy a robust indicator for localization.

\section{Conclusion}

We have introduced RT-DeepLoc, a framework that addresses the challenge of multimodal temporal deepfake localization under weak supervision by mining forgery traces from reconstruction errors. Unlike conventional methods that struggle with the heterogeneity of forgeries, our approach anchors the detection process on the regularities of authentic content. Through the integration of the FDN for anomaly discovery, AICL for targeted feature separation, and MTLR for cross-stream consistency, RT-DeepLoc provides a stable and precise localization mechanism. The superior performance across large-scale datasets confirms that modeling the ``bounded" space of real data is a more effective paradigm for generalizing to unseen attacks than modeling ``unbounded" forgery patterns. We hope this work inspires further research into leveraging self-supervised generative priors for robust and fine-grained deepfake analysis.


\bibliographystyle{ACM-Reference-Format}
\bibliography{main}

\appendix

\end{document}